# Natural Language Processing for Drug Discovery Knowledge Graphs: promises and pitfalls

J. Charles G. Jeynes[1], Tim James[1], Matthew Corney[1]

[1]Evotec (UK) Ltd., *in silico* Research and Development, 114 Innovation Drive, Milton Park, Abingdon, Oxfordshire OX14 4RZ, United Kingdom;

**Author Information**

Corresponding Authors
*E-mail: charlie.jeynes@evotec.com
tim.james@evotec.com

**Keywords:** NLP, named entity recognition, named entity linking, normalization, ontologies, DrugBank, SemMedDB, PubTator, database, heterogeneous data, unstructured text

**Running head:** NLP for knowledge graphs

## Abstract
Building and analysing knowledge graphs (KGs) to aid drug discovery is a topical area of research. A salient feature of KGs is their ability to combine many heterogeneous data sources in a format that facilitates discovering connections. The utility of KGs has been exemplified in areas such as drug repurposing, with insights made through manual exploration and modelling of the data. In this article, we discuss promises and pitfalls of using natural language processing (NLP) to mine 'unstructured text' - typically from scientific literature - as a data source for KGs. This draws on our experience of initially parsing 'structured' data sources – such as ChEMBL – as the basis for data within a KG, and then enriching or expanding upon them using NLP. The fundamental promise of NLP for KGs is the automated extraction of data from millions of documents – a task practically impossible to do via human curation alone. However, there are many potential pitfalls in NLP-KG pipelines – such as incorrect named entity recognition and ontology linking – all of which could ultimately lead to erroneous inferences and conclusions.

## 1. Introduction

The explosion in biomedical data over the last few decades has led to a pressing need for a coherent framework to manage the information. This is particularly true for drug discovery where many factors influence the success or failure of a program as it passes through the various stages, from target prioritisation, hit identification, lead optimisation through to clinical trials. Knowledge graphs (1) are a promising framework for this task as heterogeneous data sources can be combined and analysed, with entities such as drugs or genes represented as 'nodes' and relationships or links between nodes represented as 'edges'.

Within the biomedical domain there are several examples of publicly available KGs. Generally, these have been constructed by parsing publicly available 'structured' (mostly manually curated) databases into a KG compatible format known as a semantic triple (subject-verb-object or subject-predicate-object) such as drug-TREATS-disease. These databases include STRING, Uniprot, ChEMBL, DrugBank and Reactome, all of which focus on a particular biomedical domain (see (2) and (3) for reviews of biomedical databases relevant to constructing KGs). Examples of publicly available KGs include CROssBAR (4), SPOKE (which includes Hetionet (5)) and Cornell Universities' KG (6). These KGs have comparable but different 'schemas' (how nodes and edges are named and arranged) and incorporate a different number of underlying datasets. Also, biological entities (nodes) can be normalised to equally valid but different ontologies (e.g. ENTREZ (7) vs ENSEMBL (8) gene IDs) in comparable KGs, making direct comparisons non-trivial. Strikingly, none of these KGs include arguably the largest data source available – unstructured text in scientific articles like journals, patents, and clinical notes. That said, there are examples of KGs that combine NLP-derived triples with those from structured databases. For instance, the majority of triples (~80%) in AstraZeneca's "Biomedical Insights Knowledge Graph" are derived from their NLP pipelines, with the rest of the data coming from 39 structured datasets (9). Other companies that have built KGs combining data from their NLP pipelines and public databases include, Euretos, Tellic (10) and Benevolent AI (11).

To our knowledge, there are very few articles that directly investigate the expansion of biomedical knowledge graphs using NLP. In one example, Nicholson *et al.* compared their NLP approach of extracting associations of biomedical entities from sentences in the scientific literature (12) to Hetionet v1 (5), which is a biomedical KG. They compared four data subsets, including "disease-ASSOCIATED-gene" and "compound-TREATS-disease", and found that their NLP method recalled around 20-30% of the edges compared to Hetionet but added thousands of novel ones. For example, with "compound-TREATS-disease" triples, NLP recalled 30% of existing edges from Hetionet but added

6,282 new ones. Their NLP methodology showed an AUROC of around 0.6-0.85 depending on the type of triple, meaning that a relatively large proportion of the novel finds are wrong in some way. Nevertheless, they concluded that NLP could incorporate novel triples into their source KG (Hetionet).

In this article, we highlight some examples of how we are using NLP to extract information from unstructured scientific text to expand Evotec's KG. We currently use commercial NLP software to create rules-based queries that extract specific information from text, which can be then parsed into the KG. This article is not meant to be a description of Evotec's knowledge graph, however, nor an explanation of our NLP pipelines. How Evotec's KG is constructed and queried will be the subject of another article.

To highlight some pitfalls, we have used examples from the Semantic Medline database (SemMedDB) (13) and PubTator (14) databases. However, it is worth noting that the examples picked are purely illustrative of the variety of issues that must be considered when supplementing structured data sources with NLP-derived insights. In general, both databases are highly valuable with excellent accompanying tools and research.

As this article is in an edition focusing on "High Performance Computing" (HPC) it is worth noting that all the results and examples explained herein would simply not be practically possible without HPC technology. NLP pipelines are generally "embarrassingly parallel" and optimised to return results from millions of documents within seconds. The technicalities of how NLP algorithms and pipelines are constructed and optimised is an active area of research. However, it is not the focus of this article, where instead we discuss the results that certain current NLP implementations can offer and possible considerations to ensure the greatest post pipeline data quality.

## 2. Promises

The essential promise of NLP is the automated coverage of millions of documents from many sources such as scientific literature, clinician notes, and patents. In the next few sections, we highlight some of these promises with specific examples drawn from Evotec's own biomedical knowledge graph.

### 2.1 Relationship verbs and/or causality between entities

#### 2.1.1 Example: protein-protein interactions

The following example of a protein-protein interaction highlights how NLP-derived data can provide additional information and extra insights compared to available structured data sources in this domain.

Figure 1 illustrates the relationship between two proteins; MAP2K1 and MAPK3. These two proteins play a part in the MAPK signaling pathway, which is important in controlling cell division and is often aberrated in cancers (15) . Data from STRING (a structured data source (16)) shows that these two proteins physically bind to each other - data in STRING is derived from biochemical experiments and co-occurrences in the literature (amongst other measures), but generally there is little information describing *how* proteins affect each other. We have used NLP to expand the relationship (and many other protein relationships like it) to extract verbs linking the two proteins. In this case, we have evidence from NLP that MAP2K1 phosphorylates and activates MAPK3 from sentences such as: "Mek1/2 are direct substrates of Raf kinases and phosphorylated Mek1/2 activate downstream Erk1/2." (MEK2 = MAP2K2, ERK2 = MAPK1) (17). Understanding the mechanistic relationship between proteins is crucial in many aspects of drug discovery. For instance, if we wanted to indirectly inhibit the activity of MAPK3 (perhaps because it had proven difficult to drug directly), we could infer from this NLP information that inhibiting MAP2K1 could be an avenue to pursue. Such an inference would be much harder to make without the causative directionality added by the NLP.

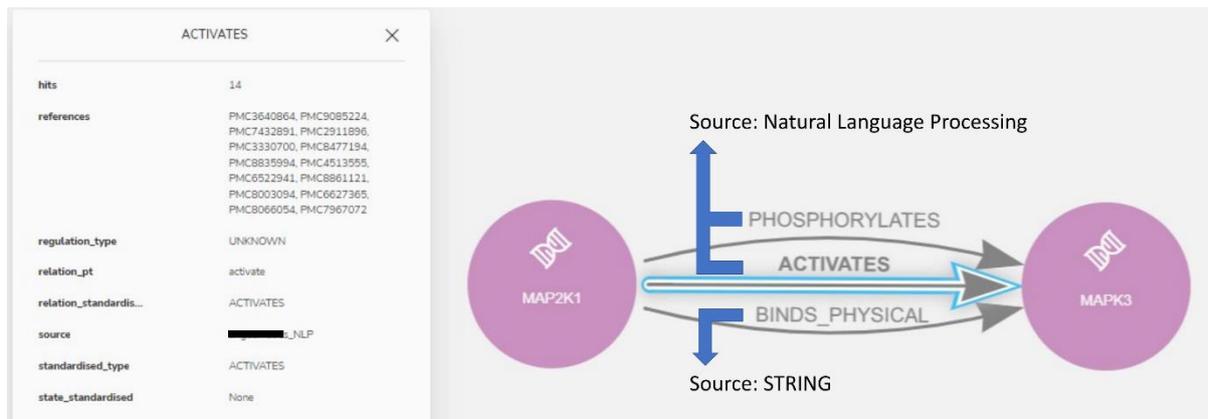

*Figure 1: An example of protein-protein interaction data in Evotec's knowledge graph. This example illustrates how triples (subject-predicate-object) extracted by NLP add valuable information (like the verbs "activates" and "phosphorylates") to structured data (here from the STRING protein-protein interaction database). From the left-hand panel titled "ACTIVATES" it can be seen that there are 14 articles ("hits" and "references") that contain phrases that equate to "MAP2K1 activates MAPK3". Future work involves displaying sentences and links to the user.*

### 2.1.2 Case study: Prioritising protein targets based on their association with a specific protein, cancer and/or arthritis

In this section, we illustrate the promise that knowledge graphs hold, exemplified by a project at Evotec. In this project, an experiment had identified a number of proteins that were associated with the regulation of a specific protein (for confidentiality we shall call this proteinX), with the caveat that there were likely to be many false positives in this list due to experimental uncertainty. From this list of proteins, we wanted to understand the plausibility of their involvement in regulation of proteinX, but also which had associations with cancer and/or arthritis. Ranking the proteins for relevance was highly desirable as further experimentation was costly and only one or two candidates could feasibly be studied in detail.

Figure 2 shows a network that was created to analyse the associations of the list of proteins. As part of our analysis, we used a "betweenness centrality" algorithm (18) on the network to indicate which proteins were most connected in terms of shortest paths between pairs of nodes. The results of this algorithm are incorporated into Figure 2, where protein nodes with a higher "betweenness centrality" score are larger. Other algorithms such as VoteRank (19) were used for different perspectives on the ranking of the proteins. We made several observations from this network. Firstly, we can see that there is a group of six proteins that have no links to any other entity in the network; these are probably false positives from the experiment. Secondly, proteinX is only directly linked to three other proteins. Thirdly, one of the directly connected proteins is a "hub" protein that has the highest "betweeness centrality" score in the network. Fourthly, the thick line between the 'highly connected protein' and 'proteinX' indicate that that this relationship is highly annotated by way of NLP; closer inspection shows there is strong evidence from the literature that the 'highly connected protein' ACTIVATES 'proteinX'. Overall, out of the initial list of proteins, at least one - the 'highly connected protein' - has a strong association with proteinX, cancer and arthritis and so might be a suitable candidate to take forward for further experiments.

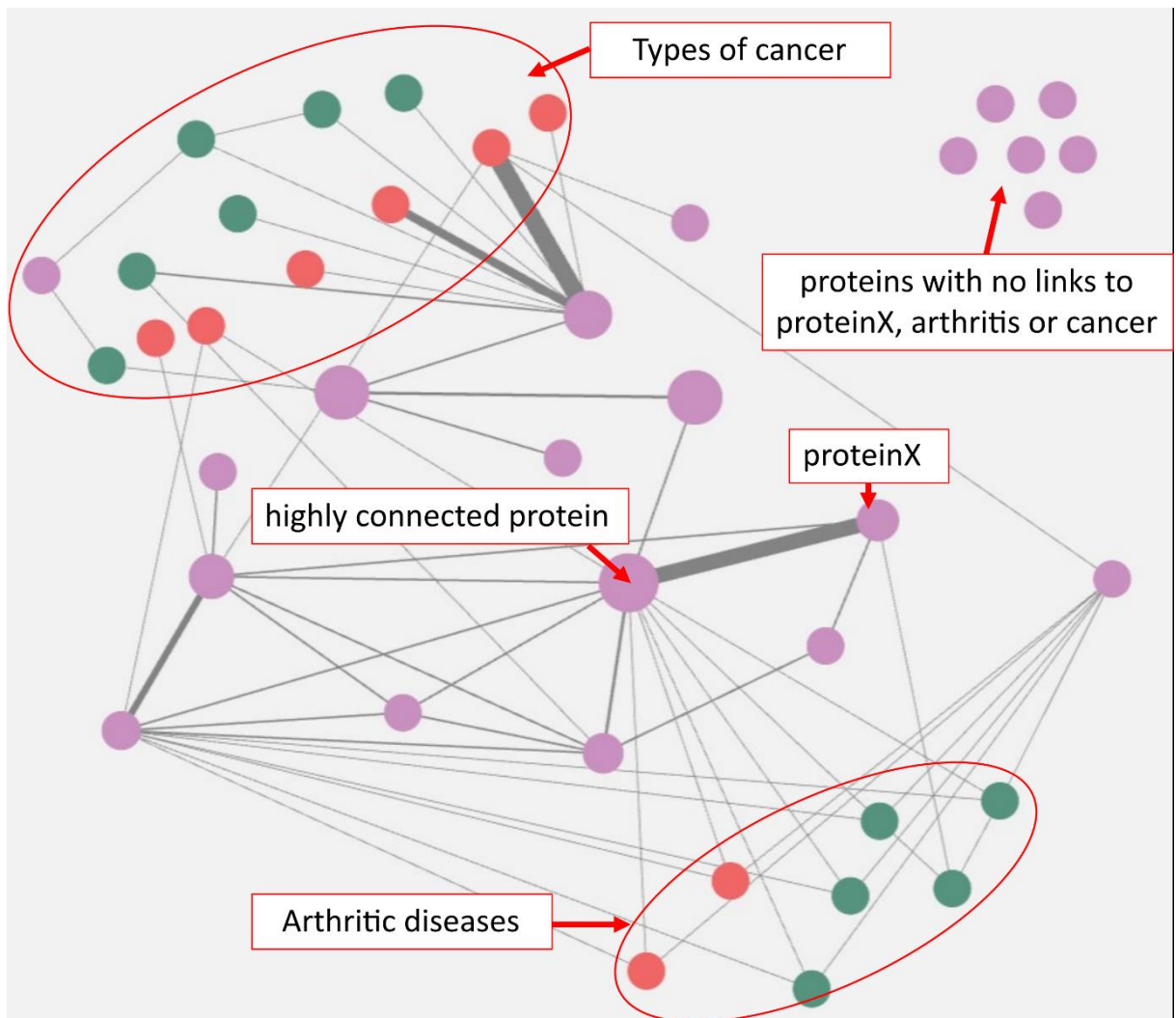

*Figure 2: An example of a mini-network or sub-graph made using data from Evotec's KG, focusing on protein-protein interactions between a specific list of proteins and their connections with cancer or arthritis. The motivation was to explore which proteins were likely regulators of 'proteinX', but were also connected to cancer and/or arthritis. The purple nodes represent proteins, with the node size scaled to reflect their 'betweenesss centrality' scores. The thickness of the network edges reflects the amount of evidence supporting the corresponding relationship. One protein (here labelled 'highly connected protein') might be a suitable candidate for further experimental testing as it has the highest score and substantial NLP-derived evidence that it activates 'proteinX'.*

## 2.2 Other examples of expanding KGs using NLP

### 2.2.1 Example: drug-TREATS-disease

In this section, we discuss how evidence for a particularly important relationship in a biomedical knowledge graph – if a drug treats or is associated with a disease - is expanded using NLP in Evotec's KG.

In the very strictest terms, the triple 'drug-TREATS-disease' should arguably be defined by data sources such as the FDA approved drugs list (*Approved Drug Products with Therapeutic Equivalence Evaluations | Orange Book*), where a drug is officially used as a treatment for a disease. However, this is a relatively small dataset compared to the many meta-analyses and trials that associate a given drug with a disease. This additional data could be very useful when investigating tasks such as drug repurposing. In the following, we argue that NLP can be used to extract useful 'drug-IS_ASSOCIATED-disease' triples. However, without due care, NLP can also pollute 'drug-TREATS-disease' data in a KG with misinformation.

Figure 3 shows a Venn diagram comparing three data sources from which we have extracted 'drug-TREATS-disease' triples. These data sources are:

1. DrugBank – a structured database (21) that is considered a 'gold-standard' because it has been manually curated from several sources including FDA-approved drugs. We extracted all drugs that had an indication for a disease.

2. A 'strict NLP query' that we applied to the entirety of MEDLINE using commercial NLP software. The query extracted all proteins and diseases that were within one word of 'treat*' (* wildcard term for derivatives like 'treatment', treated, etc) and within twenty words of 'conclusion*'.

3. The subset 'subject-TREATS-object' of the SemMed database (13).

To allow for comparison, each entity in the datasets were mapped to a Unified Medical Language System (UMLS) Concept Unique Identifier (CUI) for normalization (22). We avoided comparing the raw text of entities as this can be misleading due to the occurrence of multiple synonyms.

In Figure 3, it can be seen that there are 9,742 unique triples from the DrugBank dataset, such as 'carpecitabine-TREATS-malignant neoplasm of the fallopian tubes'. In comparison, from our 'strict NLP query' there are 21,099 unique triples, like 'mycophenolate-TREATS-birdshot chorioretinopathy'.

We manually inspected several of these triples from the 'strict NLP query' and many came from the conclusion section of a retrospective or meta-analysis abstract. One example - 'mycophenolate-TREATS-birdshot chorioretinopathy' - is drawn from sentences such as; 'Conclusions: Derivatives of mycophenolic acid are effective and safe drugs for the treatment of BSCR [birdshot chorioretinopathy]' (23).

Interestingly, there are only 54 triples that intersect with the DrugBank dataset. One reason why there are few NLP-derived triples that overlap with DrugBank could be because NLP tends to find sentences in retrospective or meta-analysis abstracts. This contrasts with DrugBank which takes sources such as the FDA drug listings as its primary data. It is expected that we will find many drugs associated with a disease from the literature that do not make it through to clinical application.

However, it is surprising that we find so few DrugBank triples using NLP, as we might have expected there to have been many mentions of a drug with a disease in the literature prior to it becoming a treatment in the clinic. The scarce overlap is probably due to the specificity of the disease that the drug is eventually indicated for and the 'strictness' of our NLP query not recalling all the possible combinations of the drug with a specific disease. So, for instance, in DrugBank the drug "Capecitabine" is indicated for several cancers including "malignant neoplasm of the fallopian tube [CUI:C0153579]". Similarly, from our "strict NLP query" the same drug is associated with over twenty types of cancer but not that specific cancer ("malignant neoplasm of the fallopian tube"). Because there are hundreds of different types of specific cancers each with their own CUI, and we are comparing the intersection using these CUIs, it becomes less surprising that there is little overlap. A combination of the 'strictness' of our query missing information from the literature and/or the evolution of the drugs' use in terms of disease specificity from preclinical to a clinical setting could be an explanation.

For many KG tasks like link prediction or other hypothesis generation we argue that triples from the 'strict NLP query' add extra useful information. Appropriate tagging of NLP-derived triples is, however, critical to enable easy identification and filtering. For example, we use 'drug-IS_ASSOCIATED-disease' rather than the more specific 'drug-TREATS-disease' assertion, which may be an important distinction. It is likely that most of these novel triples are sufficiently accurate to provide new insights into tasks like drug repurposing. Such insights would be much harder to attain if the NLP data was omitted from the knowledge graph.

In contrast, the data in SemMedDB exemplifies many of the potential pitfalls associated with adding NLP-derived data to KGs. Briefly – as this is described in more detail in section 3 – from Figure 3 it can be seen that SemMedDB has 188,662 unique 'drug-TREATS-disease' triples. However, many of these are unhelpful or misleading in some way. Around 40,000 of the 'drug-TREATS-disease' triples have a 'drug' which is in fact a drug classification rather than a specific agent – for example 'vasoactive agents-TREATS-shock'.

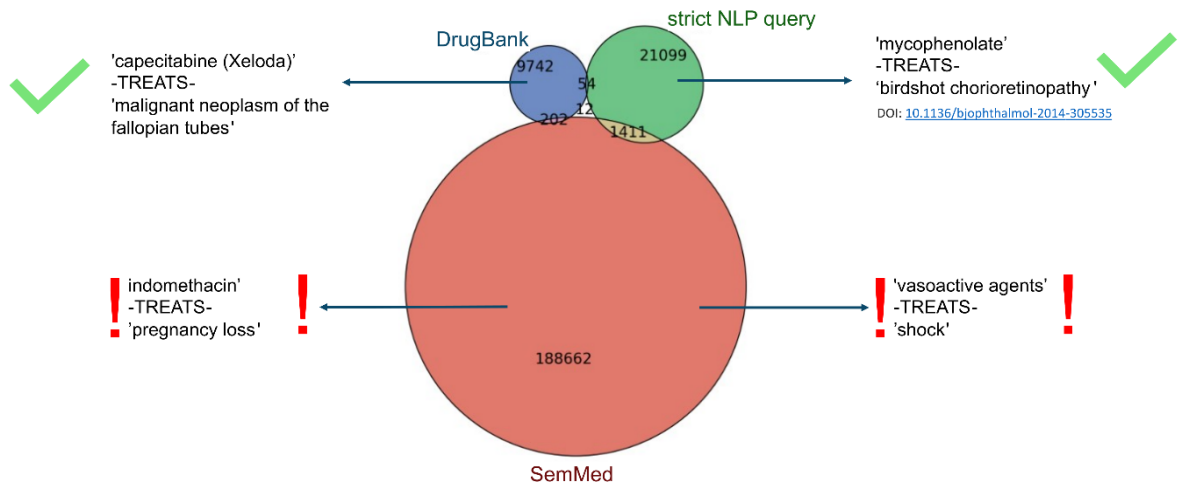

Figure 3: A Venn diagram comparing 'drug-TREATS-disease' triples extracted from three sources; 1. DrugBank, 2. a 'strict NLP query' that we constructed and used to extract information from the entirety of MEDLINE (~33 million abstracts) and 3. SemMedDB. The numbers show how many unique triples each of the datasets have and how many intersect with the other datasets. The text shows examples from each of the datasets with ticks indicating that we felt the triple represented a true reflection of the source material or sentence, while exclamation marks indicate triples that were wrong or misleading (see section 3.2 for more discussion on these).

### 2.2.2 Example: gene-HAS_FEATURE-ProteinFeature

NLP has the promise of adding crucial contextual information so that a KG is ultimately more accurate and useful. We demonstrate this using the "gene-HAS_FEATURE-ProteinFeature" triple that we have included in the Evotec KG.

Figure 4a shows the record for the AHI1 gene, where the majority of ProteinFeatures have been extracted from the literature using our NLP pipelines (other sources of information for ProteinFeatures include UniProt). One ProteinFeature is highlighted as an example; here the variant T304fs*309 is reported to be associated with mutagenesis. This information was extracted and normalised from the sentence shown in the field 'text_mentions' from the article shown in 'text_references'.

This example is closely related to a pitfall described later (see section 3.2). There we describe how the NLP-derived triple 'TP53 gene-CAUSES-neuroblastoma' is inaccurate because it is the *dysregulation* of this gene that can lead to diseases, not the gene itself. With Evotec's KG schema, we mitigate against this sort of error by including ProteinFeature nodes that can, in turn, be linked to diseases. Figure 4b shows an example of this, with:

"TP53 -> HAS_FEATURE -> TP53_gene mutation -> MUTATION_LINKED_TO -> Keratosis".

Here, the "TP53 gene mutation" is the ProteinFeature. The 'Relationship properties' box shows the evidence for this link in the sentence (see "text_mentions"), the article ("text_reference") and that NLP was the source of the evidence ("source"). Using this schema, it is clear to the user that it a mutation of TP53 that is associated with keratosis.

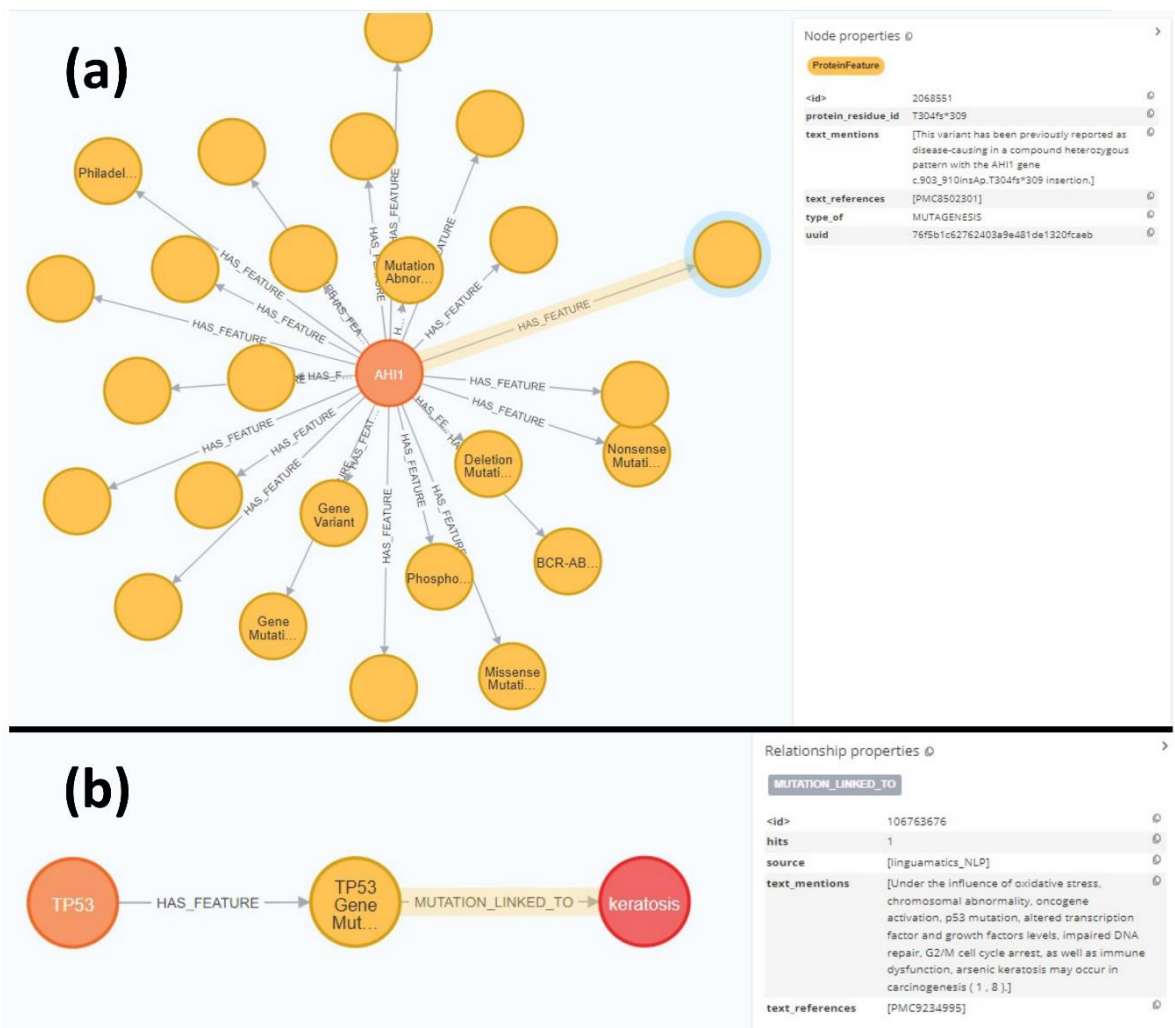

*Figure 4: (a) An example of how NLP can add contextual information that is crucial for biological interpretation. In Evotec's KG, proteins can have links to a node called 'ProteinFeature' that has attributes such as 'Mutagenesis'. These attributes have been extracted from sentences in the scientific literature (as shown in the field 'text_mentions') or from other sources such as Uniprot. (b) Protein features can be linked to diseases. Here we show an example of a mutation in the TP53 gene that is associated with the disease keratosis – a link derived from the sentence shown in the 'text_mentions' box. This example is closely related to a pitfall in using NLP to construct KGs, where genes can be linked to diseases but omitting the context, such as a mutation (see section 3.2).*

## 2.2.3 Inclusion of data sources such as Electronic Health Records (EHR)

Besides the scientific literature, there are many other sources of unstructured texted - including Electronic Health Records (EHR) and patents - which hold much promise for mining using NLP.

EHRs are important as they capture details such as drug administration regimes, comorbidities, and side-effects within day-to-day practice in the clinic. They can be particularly challenging for NLP algorithms for several reasons including difficulty identifying entities because of colloquialism and jargon used within the clinical profession not well captured by existing ontologies. That said, there are a number of papers on the topic of NLP, knowledge graphs and EHRs. These include Finlayson *et al.* who computed the co-occurrence of one million clinical concepts from the raw text of 20 million clinical notes spanning 19 years from the STRIDE dataset (24,25). There are a number of other freely available and anonymised EHR datasources including eICU (26), MIMIC-III (27) and UK Biobank (28). In the context of Evotec's KG, adding information extracted via NLP from EHRs could expand data from the SIDER ('Side Effect Resource', http://sideeffects.embl.de/) database of drugs and adverse drug reactions (29). Further, information could be extracted from EHR which are not well covered by existing databases such as disease comorbidities (30).

## 3. Pitfalls

### 3.1 Entities are incorrectly identified leading to erroneous relationships.

Accurate NLP in biomedicine generally relies on robust Named Entity Recognition (NER) and entity linking to ontologies such as UMLS (31). These ontologies contain information about the synonyms of various entities such as genes, chemicals and diseases. However, because of the myriad acronyms and synonyms that exist within the biomedical domain, the task of accurate NER is challenging and fraught with pitfalls.

Figure 5 is an illustrative example showing how an incorrectly identified protein could lead to an erroneous triple being added to a knowledge graph. Here, PubTator (14) – a publicly available dataset made possible using various NER engines including Tagger One (32) – has been used to search for 'COX1' and 'aspirin'. Unfortunately, 'COX1' is a synonym for two completely different proteins; 'cytochrome c oxidase 1' (Uniprot accession number P00395) and 'cyclo-oxygenase 1' (aka prostaglandin synthase I, Uniprot accession number P23219). In this instance, PubTator has normalised 'COX1' to 'cytochrome c oxidase 1', which in this context is an incorrect assignment.

This is important as PubTator has been used as the underlying data for several knowledge graphs. For example, (33) built a KG from a dataset called the Global Network of Biomedical Relationships (34), which in turn used PubTator as the basis for its entity recognition. It is likely that many downstream tasks in KG analysis such as link prediction are less accurate because of these sorts of errors.

There are many other ambiguous biomedical acronyms. For example, 'TTF-1' ('thyroid transcription factor 1') can be confused with 'TTF1' ('transcription termination factor 1'), which are two distinct proteins (35). In a wider medical setting, 'RA' can stand for 'Right Atrium', 'Rheumatoid Arthritis', or 'Room Air' depending on the context.

The prevalence of NER errors, such as that illustrated in Figure 5, is likely dependent on the NLP tool used. Authors employing state-of-the-art neural network models claim disambiguation accuracy >0.8 (36). The commercial NLP software we currently use at Evotec has a 'disambiguation score', which can be tuned for precision versus recall. Usually, when we run our queries we set the disambiguation threshold high to ensure that the entities returned are accurate. There are occasions where the 'hits' returned for a particular entity are so few or ambiguous in nature that they warrant the high recall returned by using a lower disambiguation threshold. Typically, we would then manually check the results for false positives and filter the data appropriately.

*Figure 5: An example of a named entity recognition (NER) error that could have serious implications if such data was included in a knowledge graph. Here, the mention 'COX1' in the context of 'aspirin' has been incorrectly identified as the gene 'cytochrome c oxidase subunit I' in PubTator. The correct gene association with aspirin in this context is 'cyclo-oxygenase 1' otherwise known as Prostaglandin G/H synthase 1 (Uniprot accession number P23219). The screenshot is taken from the PubTator user interface (https://www.ncbi.nlm.nih.gov/research/pubtator/).*

### 3.2 Relationships are wrong because they lack context

This example comes from the SemMedDB database (13), which uses a NER engine called 'meta-mapper' to identify biomedical entities and a subject-predicate-object extraction tool called 'SemRep' (37). The entirety of MEDLINE is parsed to create the SemMedDB dataset. Figure 6 shows an example of one such triple that is, at best, highly misleading because crucial context is omitted. The bare triple 'TP53 gene-CAUSES-Neuroblastoma' is incorrect as TP53 regulates the cell cycle and so – when functioning correctly – prevents cancer rather than causes it (38). In the sentence from which the triple is extracted, the key context word is 'dysregulation'. Thus, a more accurate representation of the sentence would be 'TP53[dysregulated]-CAUSES-Neuroblastoma'.

Drawing conclusions from networks with this sort of error could lead to bad decisions. Taken literally, if 'TP53 causes neuroblastoma' a reasonable hypothesis might be that inhibiting TP53 could treat the disease. In reality, the reverse is true; if we were to choose TP53 as a target to treat neuroblastoma we would probably want to restore its wild-type function rather than inhibiting it. The molecular biology of TP53 is well studied so this error is easily identified. However, there are many proteins whose functions are unclear and where this type of error could easily go unnoticed.

To mitigate against the pitfall raised above, Evotec's knowledge graph has a "ProteinFeature" node that allows us to assess more accurately which features of a protein are related to diseases, whether that feature might be a mutation or a modification. Here we refer readers back to section 2.2.2 for a detailed example.

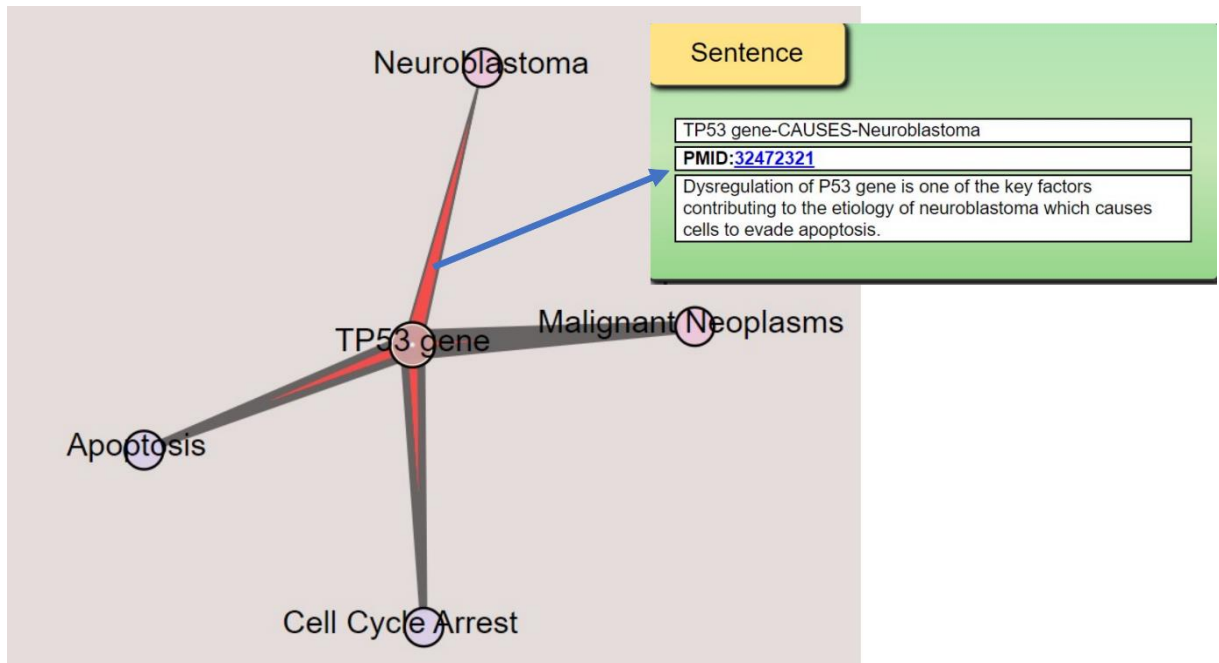

*Figure 6: An illustration of how assertions derived from NLP can be wrong because context is missing. Here, SemMedDB asserts that 'TP53 gene-CAUSES-Neuroblastoma', whereas, in reality, it is the dysregulation of TP53 that is a factor causing neuroblastoma. The same error also occurs with 'Malignant Neoplasms'. The screenshot is from the SemMedDB browser after searching for 'TP53' and filtering on 'CAUSES' (https://ii.nlm.nih.gov/SemMed/semmed.html - a UMLS licence is required).*

### 3.3 Adding noise (assertions are not incorrect but are generally unhelpful due to insufficient granularity)

Another pitfall is adding 'noise' to the knowledge graph, which we define as data that is not necessarily incorrect but just generally unhelpful. Figure 7 shows an example of this from SemMedDB. Here, we have the triple 'Vasoactive agent-TREATS-shock', which is derived from sentences such as 'We present a comprehensive review of conventional, rescue and novel vasoactive agents including their pharmacology and evidence supporting their use in vasodilatory shock'. The entity 'vasoactive agent' is a legitimate biological entity with an associated UMLS CUI (22), which is why it has been identified by SemMed's NER algorithm. However, we argue that for most, if not all, knowledge graph tasks these so-called 'parent terms' (other examples of which include 'pharmaceutical substances', 'vaccines' and 'antibiotic agents') in an ontology like UMLS are just adding noise. Almost always, we are interested in which *specific* vasoactive agents treat a disease. Fortunately, it is relatively simple to filter out these terms by manually compiling a list – made easier as many nuisance terms often contain keywords such as 'agent' – or by filtering using the hierarchical structure of the ontology. Thus, SemMed can be filtered for more specific triples. Of note, we use commercial NLP software that has a 'leaf node only' option, so that 'parent terms' can be easily excluded.

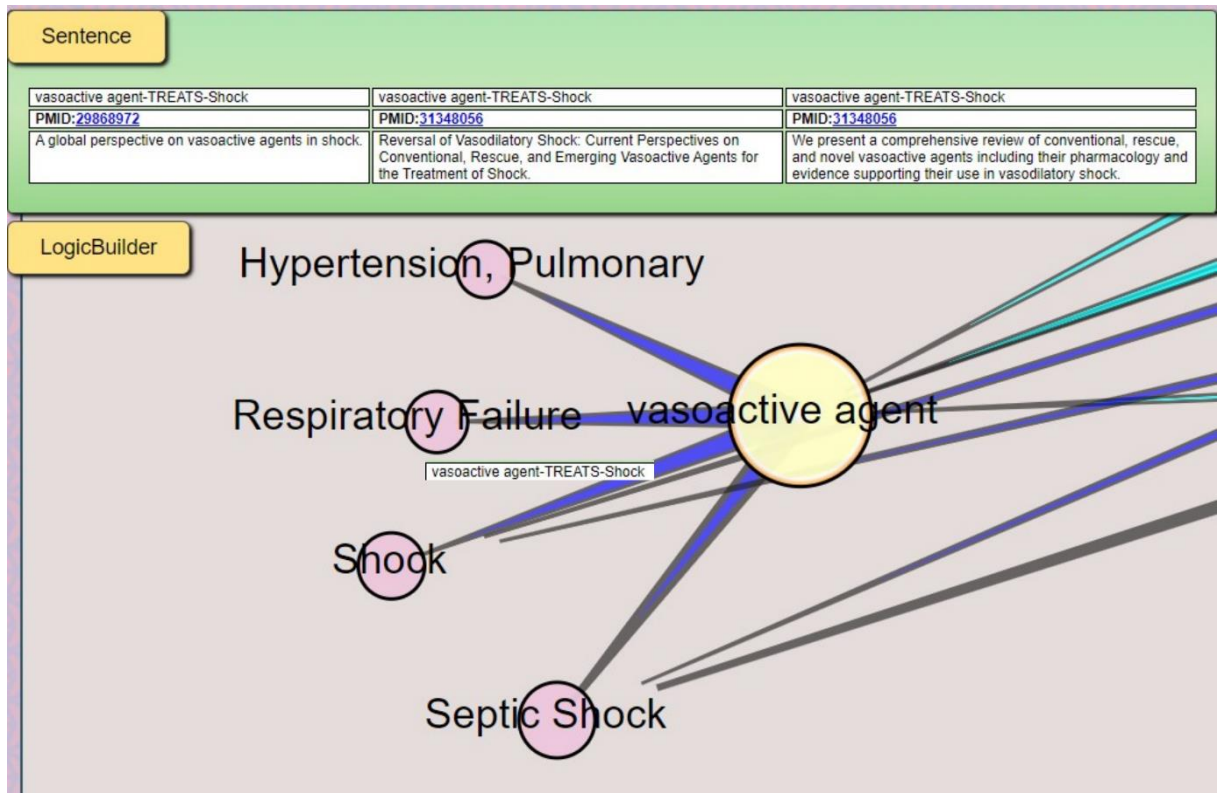

*Figure 7: An illustration of how NLP can derive assertions that can be considered 'noise' – that is, not untrue but arguably unhelpful. This example from SemMedDB asserts the generic triple 'vasoactive agent-TREATS-shock' (and also hypertension, respiratory failure and septic shock). While the assertion is undoubtably true, the term 'vasoactive agent' is so general as to be unhelpful for most drug discovery purposes. The screenshot is from the SemMedDB browser after searching for 'vasoactive agent' and filtering on 'TREATS' (https://ii.nlm.nih.gov/SemMed/semmed.html - a UMLS licence is required)*

### 3.4 Misrepresenting certainty of assertion

Another pitfall is where a triple extracted by NLP over-simplifies the sentence from which it came and therefore misrepresents the certainty of the assertion. Many sentences contain hypotheses, speculations, opinions or counter-factuals (e.g. drug does **not** treat disease). Without careful mitigation it is possible to include many triples in a KG derived via NLP that are hypotheses and should not be represented as facts.

We illustrate this point with an example described by Kilicoglu *et al.* (39) showing how SemMed processes sentences. The sentence '*Whether decreased VCAM-1 expression is responsible for the observed reduction in microalbuminuria, deserves further investigation.*' is represented as the triple: 'Vascular Cell Adhesion Molecule 1-DISRUPTS- Microalbuminuria'. This is clearly a misrepresentation of the sentence.

There are several strategies to account for what Kilicoglu *et al.* describe as assessing 'factuality' of assertions, ranging from rules-based trigger word identification for phrases like 'may', 'not', 'could be', 'investigation'or 'suggests' to machine learning approaches. Kilicoglu *et al.* developed a scoring system to categorise assertions as 'certain_not', 'doubtful', 'possible', 'probable' or 'certain'.

At Evotec we are working towards incorporating such a system into our pipeline for extracting triples. Currently, we exclude sentences that contain trigger words like 'not' or 'investigated'. This serves for the short term as it ensures we have high precision (certainty of an assertion) but it does also mean we potentially decrease recall. One could envisage including all assertions into a knowledge graph with a 'factuality' score, which could be filtered depending on the users' need.

## 4. Discussion

In this article, we have provided several examples of where we believe NLP holds the most promise in terms of expansion of KGs. At the same time, we have illustrated a number of pitfalls that can easily be encountered if one is not cautious when using NLP.

We have used NLP to expand Evotec's knowledge graph in an iterative and evolving process and will continue to do so as the need for new data becomes apparent. These needs are identified by using graph to try to answer real-world questions that arise in the projects we are working on, as well as building up robust datasets for AI/ML based approaches.

This workflow is illustrated in our first example where we have expanded the data relating to protein-protein interactions. We were not entirely satisfied with the data we were able to retrieve from structured data sources such as STRING because we felt useful relationship types were missing. Using NLP we could add biologically relevant verbs like 'phosphorylates' to describe the relationships between proteins. We also found we could add directional or causative verbs like 'activates' and 'inhibits', which describe the effects one protein has on another. We have found that, when exploring relationships in the KG, having this information readily available made the graph much more useful in the analysis and decision-making process. Prioritising proteins as potential targets is an example use case.

We have discussed several pitfalls we have encountered when thinking about expanding the KG using NLP or NLP-derived datasets. For example, one might be tempted to expand a KG using the SemMedDB (13) dataset, which extracts 'subject-predicate-object' triples from the entirety of MEDLINE. However, as we have illustrated, such a 'broad-brush' NLP engine can result in many misrepresentations and errors that could corrupt a KG. That is not to say that SemMedDB is not useful or to be trusted, rather it needs careful filtering before any of its data is incorporated into a KG. Our approach at Evotec to circumnavigate many of the pitfalls we have highlighted is to create highly stringent 'domain-specific' NLP queries. These ensure that extracted data is of high quality and can be trusted (high precision), perhaps at the expense of some recall.

Unfortunately, the pitfalls faced when attempting to expand KGs using NLP are compounded by the fact that biomedical KG 'schemas' (aka metagraphs or KG model) are not standardized within the field. Published biomedical KGs vary in the data they use and the 'schema' they choose to represent the data. This is particularly relevant to NLP as the possibilities of including various aspects and combinations of entities and relationships are numerous, with a corresponding potential for variation between knowledge graphs. For example, we have used the schema "Gene-HAS_FEATURE-ProteinFeature", where attributes of the ProteinFeature node have fields such as 'type_of: mutagenesis' that can then be linked to a disease. However, it could be equally valid to use a "Gene-MUTANT_VERSION-Disease" representation, with the details of the mutation being an attribute of the mutant version edge.

It is possible that in the coming years there may be a coalescence around a standardised method of constructing, analysing and interpreting a biomedical KG from publically available data sources. This would include data extracted from the literature using NLP pipelines. Standardisation efforts are already underway with examples such as BIOLINK (40), where a canonical schema or model for how a KG should be constructed is proposed. That said, the task of 'reducing' complex biological data down to the basic framework of the KG – the sematic triple – is a challenging task. There may be no 'correct' model for this, as various representations could be equally valid. However, some unification in the biomedical KG field is probably needed on the basis that, currently, two separate KGs constructed from broadly the same public sources can give wildly different predictions after embedding algorithms are applied to them. For example, in a drug repurposing task for COVID-19, only one drug overlapped from the top 30 predictions made by two separate authors using their own KGs (41,42), even though similar embedding models were used. In related work, Ratajckzak *et al.* showed that the relative ranking of drugs predicted to treat SARS-CoV2 depended on the KG (Hetionet vs DRKG) and the subset of data the embedding models were trained on (43).

Related to the above example is the fact that different users often require distinct levels of detail in a particular domain. For instance, those working in proteomics may want details on gene transcript isoforms (gene->transcript->protein), while for others working in synthetic chemistry this granularity may just obscure rather than reveal. One potential route around this (and one we use with Evotec's KG) is to expose different subsets of the data to different users. As Ratajckzak *et al.* demonstrated, domain specific models are more accurate, so creating predictive models for use in a particular domain is likely necessary. Overall, we expect that through a trial-and-error process, and as more scientists work with KGs, their usefulness to the drug discovery endeavor will improve year-by-year.

## 5. Conclusion

The process of biomedical KG construction, analysis and interpretation is still a relatively new field. The promise of using NLP to expand or even form the basis of KGs for drug discovery is huge. However, there are also many pitfalls that should be considered and could, if ignored, compromise the integrity of a KG. Here, we have discussed a few examples, including how NLP can add highly valuable descriptions and context to relationships between entities. We have also demonstrated some common pitfalls that include errors relating to named entity recognition and misrepresenting the true meaning of sentences so that opinions are introduced into the KG as facts. At Evotec, we have been updating the data and modifying the schema of our KG in an iterative fashion as we use it to answer various drug discovery related questions. We have found using domain-specific NLP queries very useful in expanding the KG where data from structured sources was lacking. Overall, we believe that carefully introduced data derived from NLP should be an essential part of a biomedical KG data curation pipeline.